\documentclass[twoside,preprint,11pt]{article}

\usepackage{jmlr2e}
\usepackage{amsmath,amsfonts,bbm}
\usepackage{color}
\usepackage{graphicx}
\usepackage{wrapfig}
\usepackage{enumitem}  % for noitemsep in lists

\newcommand{\ba}{{\pmb{a}}}
\newcommand{\bs}{{\pmb{s}}}
\newcommand{\bx}{{\pmb{x}}}
\newcommand{\bth}{{\pmb{\theta}}}

\ShortHeadings{Scalable Parameter Inference for Adaptive Domain Randomization with Isaac Gym}{Antonova, Ramos, Possas and Fox}
\firstpageno{1}

\begin{document}

\title{BayesSimIG: Scalable Parameter Inference for Adaptive Domain Randomization with IsaacGym}

\author{\name Rika Antonova \email rika.antonova@stanford.edu \\
       \addr Department of Computer Science\\
       Stanford University, USA 
       \AND
       \name Fabio Ramos \email ftozetoramos@nvidia.com \\
       \addr NVIDIA, USA
       \AND
       \name Rafael Possas \email rafael.possas@sydney.edu.au \\
       \addr School of Computer Science\\
       University of Sydney, Australia
       \AND
       \name Dieter Fox \email dieterf@nvidia.com \\
       \addr NVIDIA, USA
       }

\editor{JMLR Open Source Software Track}

\hypersetup{
    bookmarks=false,        % show bookmarks bar?
    pdftitle={BayesSimIG: Scalable Parameter Inference for Adaptive Domain Randomization with IsaacGym},     % title of the document
    pdfauthor={Rika Antonova, Fabio Ramos, Rafael Possas, Dieter Fox},   % author of the document
    pdfsubject={Robotics},  % subject of the document
    pdfcreator={},   % creator of the document
    pdfproducer={}, % producer of the document
    pdfkeywords={Robotics, Reinforcement Learning, GPU-Accelerated Simulation}, % list of keywords
    colorlinks=false,       % false: boxed links; true: colored links
    linkcolor=blue,         % color of internal links (change box color with linkbordercolor)
    citecolor=black,        % color of links to bibliography
    filecolor=black,        % color of file links
    urlcolor=black          % color of external links
}

\maketitle

\begin{abstract}%   <- trailing '%' for backward compatibility of .sty file
BayesSim is a statistical technique for domain randomization in reinforcement learning based on likelihood-free inference of simulation parameters. This paper outlines BayesSimIG: a library that provides an implementation of BayesSim integrated with the recently released NVIDIA IsaacGym. This combination allows large-scale parameter inference with end-to-end GPU acceleration. Both inference and simulation get GPU speedup, with support for running more than 10K parallel simulation environments for complex robotics tasks that can have more than 100 simulation parameters to estimate. BayesSimIG provides an integration with TensorBoard to easily visualize slices of high-dimensional posteriors. The library is built in a modular way to support research experiments with novel ways to collect and process the trajectories from the parallel IsaacGym environments.
\end{abstract}


%------------------------------------------------------------------
\section{Introduction}

BayesSim is a likelihood-free inference framework for domain randomization  in reinforcement learning proposed in~\cite{ramos2019bayessim}. It allows computing flexible multimodal posteriors for simulation parameters of a complex simulator, given a dataset of simulated trajectories and a small number of observations obtained from a real system. BayesSim offers a principled way to reason about uncertainty, unlike the alternative approaches that compute point estimates of the parameters, such as various system identification methods~\citep{goodwin1977dynamic}.
In contrast to methods that fit a unimodal Gaussian, e.g.~\cite{chebotar2019closing}, BayesSim posterior is represented by a Gaussian mixture, with full covariance Gaussian components. Since Gaussian mixtures are universal approximators for densities~\citep{kostantinos2000gaussian, goodfellow2016deep}, given enough mixture components BayesSim posteriors ensure sufficient representational capacity.

BayesSim has been demonstrated to be useful for a wide range of robotics problems in simulation~\citep{mehta2020calibrating} and on hardware~\citep{barcelos2020disco, matl2020inferring, possas2020online, matl2020stressd}. Hence, the research community is interested in having an open source codebase for BayesSim. The BayesSimIG repository that we describe here offers a scalable and easy-to-use BayesSim implementation:
\setlist{nolistsep}
\begin{enumerate}[itemsep=2px]
\item[-] scalable GPU-accelerated simulation, training and inference;
\item[-] modular architecture: choices for trajectory collection  and summarization strategies;
\item[-] integration with TensorBoard for visualizing posteriors and training progress;
\item[-] examples with posteriors comprised of dozens of simulation parameters and support for larger problems with $>\!\!100$-dimensional parameter posteriors;
\item[-] ready-to-use adaptive domain randomization: full integration with the recent \mbox{IsaacGym} (IG) simulator and reinforcement learning (RL) setup~\citep{ig2021};
\item[-] support for further research directions, such as experimenting with differentiable trajectory summarizers from the signatory library~\citep{kidger2021signatory}.
\end{enumerate}
\vspace{4px}

\noindent\textbf{A brief description of the mathematical formulation: }
BayesSim starts by considering a prior $p(\bth)$ over a vector of $D$ simulation parameters $\bth = [\theta_1, ..., \theta_D]$ and a derivative-free simulator used for obtaining data/trajectories of a dynamical system. Each trajectory ${\bx}^s$ is comprised of simulated observations for states $\{\bs\}_{t=1}^T$ of a dynamical system and the controls/actions $\{\ba\}_{t=1}^T$ that were applied to the system. BayesSim then collects a few observations from the real world, e.g. a single episode/trajectory ${\bx}^r$ and uses it to compute the posterior $p\Big(\bth \Big| \big\{{\bx}_{(i)}^s\big\}_{i=1}^N, {\bx}^r\Big)$. Instead of assuming a particular form for the likelihood and estimating $p(\bx | \bth)$, BayesSim approximates the posterior by learning a conditional density $q_{\phi}(\bth | \bx)$, represented by a mixture density neural network (MDNN) with weights $\phi$. The posterior is obtained as $\hat{p}(\bth | \bx \!=\! {\bx}^r) \propto \frac{p(\bth)}{\tilde{p}(\bth)} q_{\phi}(\bth | \bx \!=\! {\bx}^r)$, which offers an option to use a proposal prior $\tilde{p}(\bth)$ used to collect simulated observations to train the conditional density.

%------------------------------------------------------------------
\section{Software Architecture}

\begin{wrapfigure}{r}{0.5\textwidth}
  \centering
  \vspace{-25px}
    \includegraphics[width=0.5\textwidth]{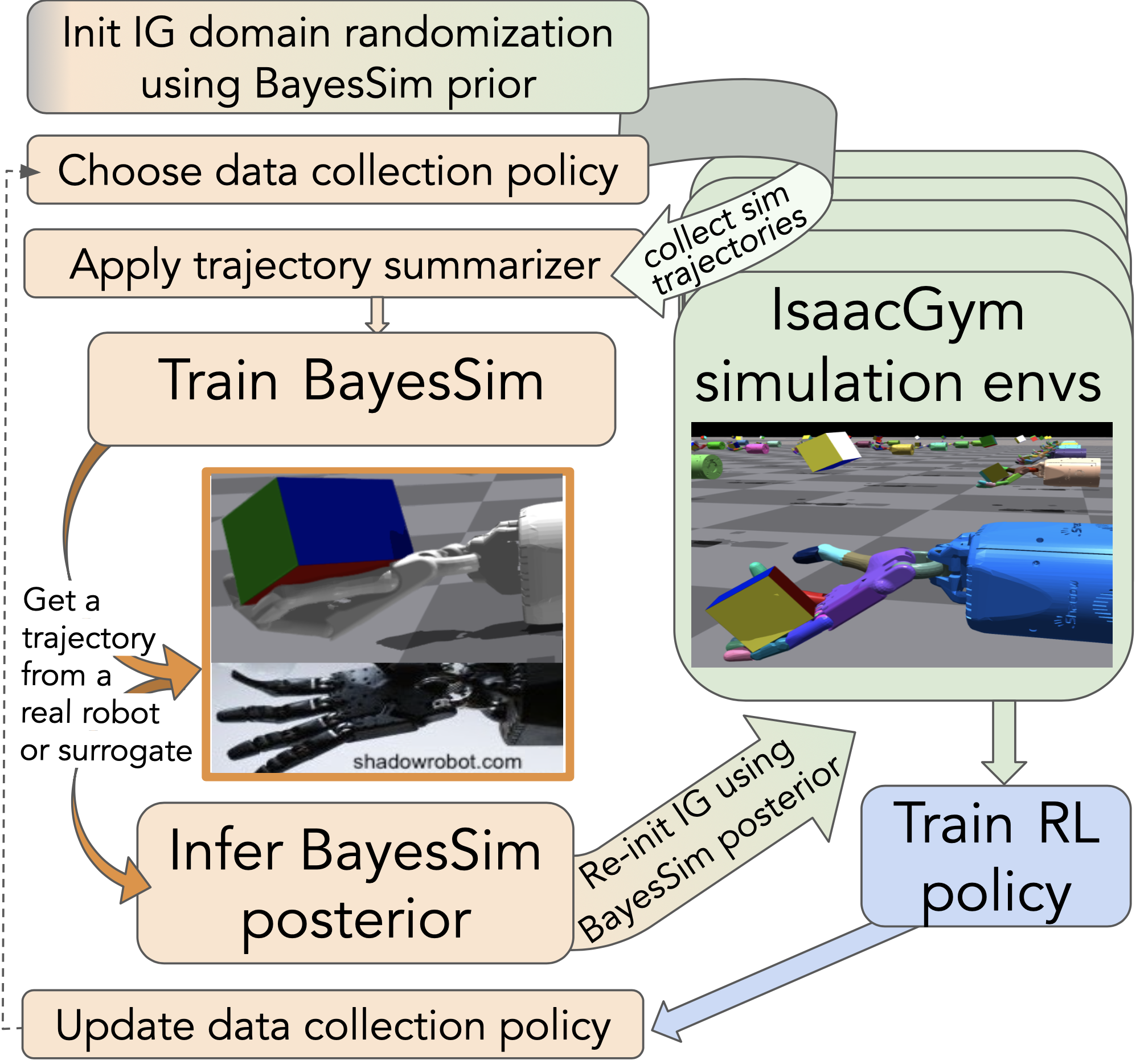}
  \vspace{-23px}
  \caption{\small{BayesSimIG main control flow.}}
  \label{fig:arch}
  \vspace{-5px}
\end{wrapfigure}

\begin{figure}[t]
  \includegraphics[width=0.58\textwidth]{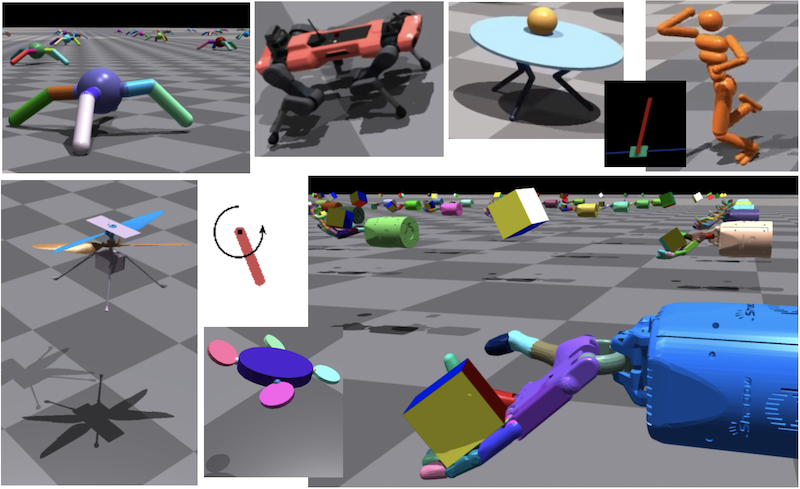}
  \includegraphics[width=0.42\textwidth]{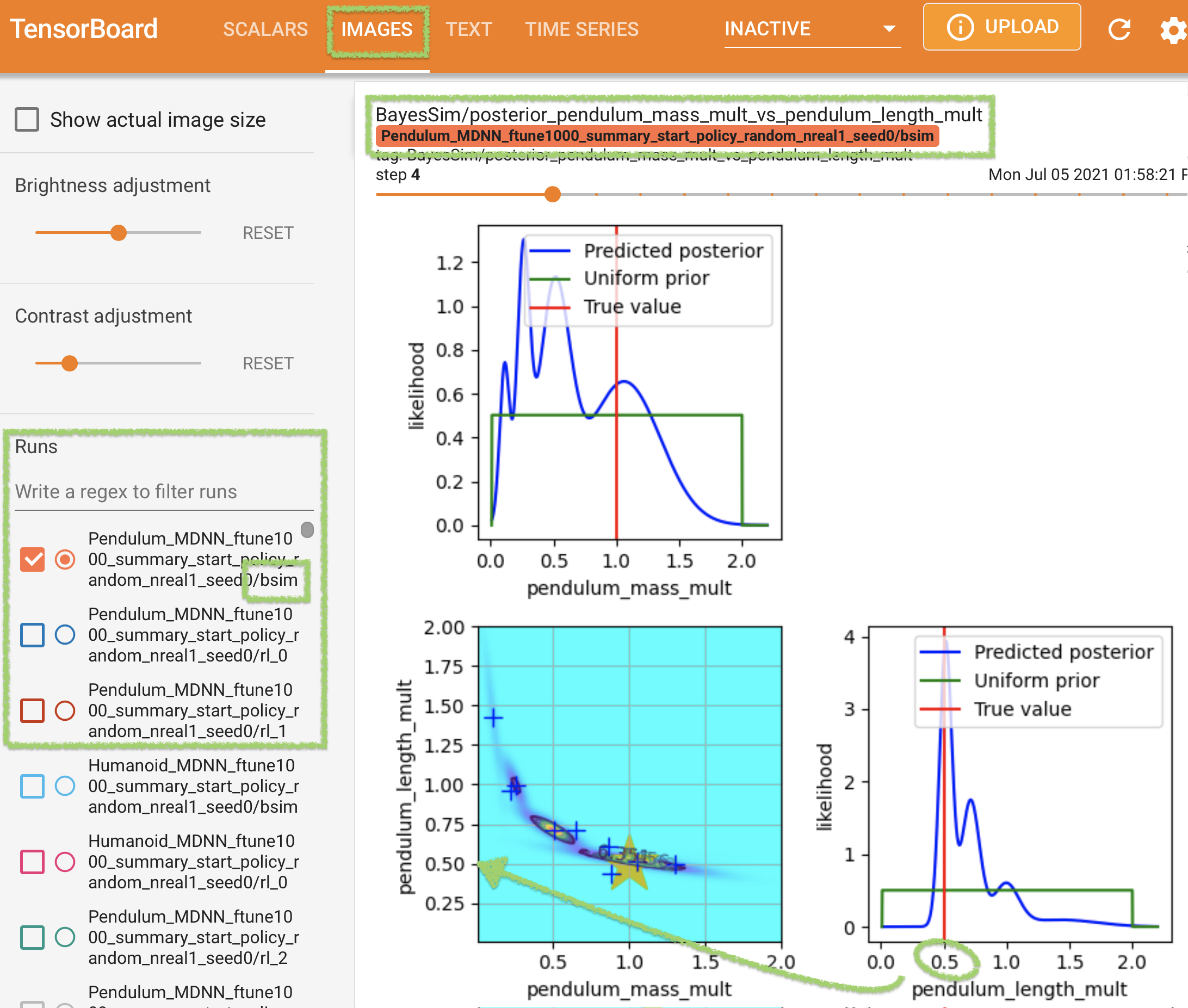}
  \vspace{-20px}
  \caption{\small{Left: IsaacGym tasks integrated with BayesSimIG: Ant, Anymal, BallBalance, Cartpole, Humanoid, Ingenuity, Pendulum, Quadcopter, ShadowHand. \\ Right: Annotated TensorBoard screenshot with RL training runs and BayesSim posterior.}}
  \label{fig:tasks_and_tb}
  \vspace{-15px}
\end{figure}

Figure~\ref{fig:arch} shows an overview of the main BayesSimIG control flow. Each component is parameterized by arguments in Yaml configuration files, for easy customization and tracking of various experiments. BayesSimIG is integrated with the environments/tasks from the recent release of~\cite{ig2021}, Figure~\ref{fig:tasks_and_tb} shows simulation visualizations.

BayesSimIG starts by reading the description of the \textbf{prior} (from the given Yaml config). Then, \textbf{data collection policy} is initialized. Users can choose from: \texttt{random, fixed, rl, rl\_randomized}; or implement a custom policy function. 

Next, we launch the parallel IG environments and collect simulated trajectories for BayesSim training. IG can easily run tens of thousands of environments/tasks on a single GPU. Before training, the trajectories are compressed by \textbf{trajectory summarizers}. Users can choose from: \texttt{start, waypoints, signatory, cross\_correlation, cross\_corr\_difference}. The original BayesSim paper used \texttt{cross\_corr\_difference}, while some follow-up work used \texttt{cross\_correlation} directly on the states, not state differences; \texttt{start} takes a short initial snippet of the trajectory; \texttt{waypoints} takes states and actions at fixed intervals. \texttt{signatory} integrates support for a recent  differentiable path signatures library~\citep{kidger2021signatory}, discussed further in Section~\ref{sec:custom_and_new}.

Next, we \textbf{train BayesSim} on the trajectory summaries to get $q_{\phi}\big(\bth \big| {\bx}_{(i)}^s\big\}_{i=1}^N\big)$. Users can select from \texttt{MDNN} or \texttt{MDRFF} models. \texttt{MDNN} is the Mixture Density Network approach~\citep{bishop1994mixture} implemented in  PyTorch with full covariance Gaussian mixture components. \texttt{MDRFF} is a Mixture Density Random Fourier Network that extracts Random Fourier Features~\citep{rahimi2007random} from trajectory summaries, achieving sharper posteriors~\citep{ramos2019bayessim}.

We are now ready to \textbf{infer the posterior from a real trajectory} ${\bx}^r$. Roboticists can obtain ${\bx}^r$ by interacting with a real robot. For the learning community we provide a surrogate `real' option. This is an IG environment/task that reads \texttt{realParams} description from the default or a user-specified config. These parameters are not known to BayesSim or other parts of the pipeline, and are used only for simulating a surrogate `real' ${\bx}^r$. 

Next, we pass the posterior to IG, so that further simulations are initialized with physics and environment properties drawn from this posterior. We then train a Reinforcement Learning (RL) agent on IG tasks with parameters randomized according to the posterior. The trained RL policy can be used for further data collection in the next BayesSimIG iteration.

BayesSim and RL networks can be either fine-tuned on each subsequent iteration, or re-initialized and trained from scratch. Users can specify the number of iterations of the whole pipeline, as well as more fine-grained arguments, such as amount of data, number of gradient updates, neural network layer sizes and activation functions.

%---------------------------------------------------------------
\section{Usage and Visualization Examples}

The first step for users is to download the IsaacGym simulator from 

\noindent \texttt{https://developer.nvidia.com/isaac-gym} and install with:
\vspace{-5px}
\begin{verbatim}
$ tar -xvzf IsaacGym.tar.gz && cd isaacgym/python/ && pip install -e .
$ export ISAACGYM_PATH=/path/to/isaacgym
\end{verbatim}
\vspace{-5px}

\noindent Users can then install and run BayesSimIG, and visualize training in TensorBoard:
\vspace{-5px}
\begin{verbatim}
$ git clone https://github.com/NVlabs/bayes-sim-ig.git
$ cd bayes-sim-ig && pip install -e .
$ python -m bayes_sim_ig.bayes_sim_main --task Pendulum  --logdir /tmp/bsim/
$ tensorboard --logdir=/tmp/bsim/ --bind_all
\end{verbatim}
\vspace{-5px}

\noindent The above commands are all that is needed to obtain a TensorBoard output similar to Figure~\ref{fig:tasks_and_tb}. For machine learning researchers interested in these advanced applications of ML methods to robotics -- no robotics background is needed to get started with BayesSimIG.
\vspace{5px}

The left sidebar in TensorBoard will show the runs with descriptive names in the format
\texttt{[Task]\_[BayesSim NN type]\_[summarizer name]\_[sampling policy]\_seed[N]}.

\begin{figure}[t]
  \includegraphics[width=1.0\textwidth]{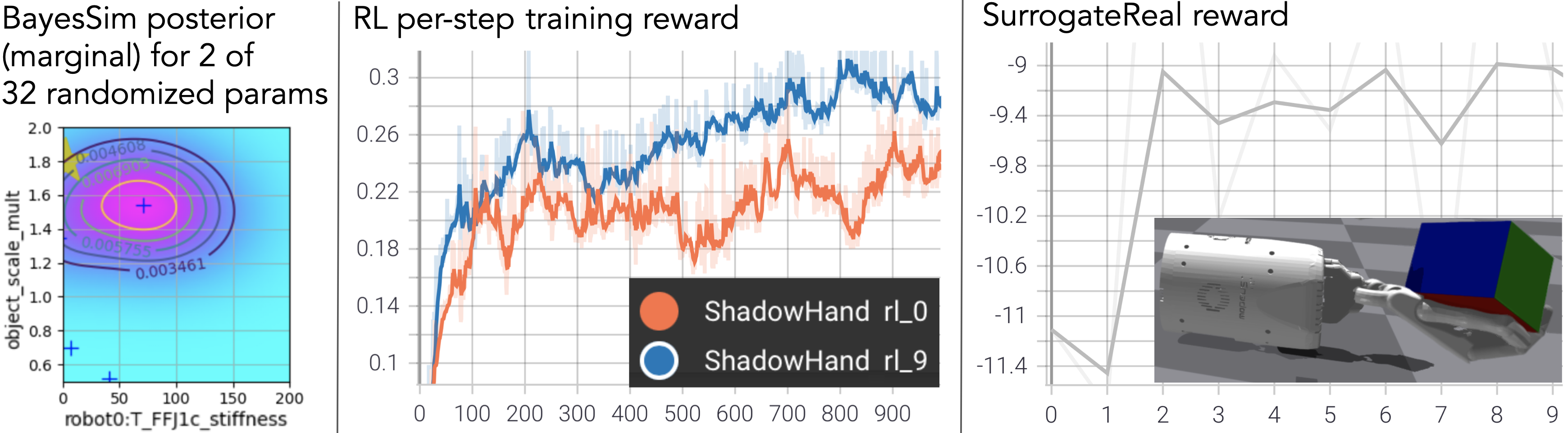}
  \vspace{-20px}
  \caption{\small{Left: a 2D slice of the posterior for the ShadowHand task (after 10 BayesSim iterations). Middle: RL reward on the 0th and 9th iterations. Right: surrogate real rewards. RL training rewards are higher than surrogate because the true object scale is large (1.8 times the default small cube), which makes the surrogate `real' task difficult. RL training includes some episodes with smaller object scales, presenting an easier problem. Still, BayesSim posterior is useful, as long as the true parameters are likely under the posterior.}}
  \label{fig:shadow_hand}
  \vspace{-10px}
\end{figure}

\noindent The \texttt{SCALARS} tab will contain plots with training/test losses and RL training statistics. Figure~\ref{fig:tasks_and_tb} shows an example posterior for the mass and length of an inverted pendulum. Figure~\ref{fig:shadow_hand} shows an example of running BayesSimIG on the ShadowHand task. This task can present an inference problem for posteriors with up to 107 dimensions. Users can consider a subset of parameters as well. Figure~\ref{fig:tasks_and_tb} shows that BayesSim posterior can speed up RL training compared to using a uniform prior. There is also an indication for the potential of better performance on a real system. 

\section{Customization and Support for New Research Directions}
\label{sec:custom_and_new}

When developing BayesSim, we integrated the ability to sample domain randomization parameters from generic distributions into the main IsaacGym codebase. The parameters that can be randomized are described in
\texttt{isaacgym/docs/rl/domainrandomization.html}. BayesSimIG also
inherits the command-line arguments from the IsaacGym RL training
launch scripts documented in \texttt{isaacgym/docs/examples/rl.html}.
The \texttt{README} in the BayesSimIG library provides more detailed instructions on customizing BayesSimIG and further example visualizations.

BayesSimIG supports experimentation with advanced ways to process trajectory data before training the core BayesSim networks. Integration with the Signatory library~\citep{kidger2021signatory} allows using differentiable path signatures as a way to summarize trajectory data. 
Path signatures (or signature transforms) allow extracting features using principled methods from path theory. These can guarantee useful properties, such as invariance under time reparameterization, and ability to extend trajectories by combining signatures (without re-computing signatures from scratch). The fact that these objects are differentiable allows backpropagating through the summarizers, making it a novel and interesting avenue for further experiments with adaptive representations for sequential data.

%---------------------------------------------
% Re-enable 4-page limit for JMLR open source track submission
% and uncomment newpage before references
%\newpage

%\appendix
%\section*{Appendix A.}

\bibliography{references}

\end{document}